\documentclass[manuscript,screen]{acmart}
\AtBeginDocument{%
  \providecommand\BibTeX{{%
    \normalfont B\kern-0.5em{\scshape i\kern-0.25em b}\kern-0.8em\TeX}}}

\setcopyright{none}

\acmConference[ACM FAcct '24]{ACM Conference on Fairness, Accountability, and Transparency}{June 03--06, 2024}{ Rio de Janeiro, Brazil}
\usepackage[ruled,vlined]{algorithm2e}

\begin{document}

\title{Counterfactual Generation with Answer Set Programming}


\author{Sopam Dasgupta}
\email{sopam.dasgupta@utdallas.edu}
\orcid{0009-0008-3594-5430}
\affiliation{%
  \institution{The University of Texas at Dallas}
  \streetaddress{800 W Campbell Rd}
  \city{Richardson}
  \state{Texas}
  \country{USA}
  \postcode{75080-3021}
}

\author{Farhad Shakerin}
\affiliation{%
  \institution{Microsoft Research}
  \country{USA}}
\email{fshakerin@microsoft.com}

\author{Joaqu\'in Arias}
\orcid{0000-0003-4148-311X}
\affiliation{%
  \institution{CETINIA, Universidad Rey Juan Carlos}
  \city{Madrid}
  \country{Spain}
}

\author{Elmer Salazar}
\email{ees101020@utdallas.edu}
\affiliation{%
  \institution{The University of Texas at Dallas}
  \streetaddress{800 W Campbell Rd}
  \city{Richardson}
  \state{Texas}
  \country{USA}
  \postcode{75080-3021}
}

\author{Gopal Gupta}
\email{gupta@utdallas.edu}
\orcid{0000-0001-9727-0362}
\affiliation{%
  \institution{The University of Texas at Dallas}
  \streetaddress{800 W Campbell Rd}
  \city{Richardson}
  \state{Texas}
  \country{USA}
  \postcode{75080-3021}
}

\renewcommand{\shortauthors}{Dasgupta, et al.}

\begin{abstract}
Machine learning models that automate decision-making are increasingly being used in consequential areas such as loan approvals, pretrial bail approval, hiring, and many more. 
Unfortunately, most of these models are black-boxes, i.e., they are unable to reveal how they reach these prediction decisions. A need for transparency demands justification for such predictions. An affected individual might also desire explanations to understand why a decision was made. Ethical and legal considerations may further require informing the individual of changes in the input attribute that could be made to produce a desirable outcome. This paper focuses on the latter problem of automatically generating   \textit{counterfactual explanations}. We propose a framework \textit{Counterfactual Generation with s(CASP) (CFGS)} that utilizes answer set programming (ASP) and the s(CASP) goal-directed ASP system to automatically generate counterfactual explanations from rules generated by \textit{rule-based machine learning (RBML)} algorithms. 
In our framework, we show how counterfactual explanations are computed and justified by imagining worlds where some or all factual assumptions are altered/changed. More importantly, we show how we can navigate between these worlds, namely, go from our original world/scenario where we obtain an undesired outcome to the imagined world/scenario where we obtain a desired/favourable outcome. 

\end{abstract}





\maketitle

\vspace{-0.05in}
\section{Introduction}

Predictive models are used in automated decision-making, for example, in the filtering process for hiring candidates for a job or approving a loan. Unfortunately, most of these models are like a black box, making understanding the reasoning behind a decision difficult. In addition, many of the decisions such models make have consequences for humans affected by them. When subject to unfavorable decisions, judgments, or outcomes, humans desire a satisfactory explanation. This desire for transparency is essential, regardless of whether an automated system (e.g., a data-driven prediction model) or other humans make such consequential decisions. Hence, making such consequential decisions explainable to people is a challenge. For a decision made by a machine learning system, Wachter et al. \cite{wachter} highlight an approach where a counterfactual is generated to explain the reasoning behind a decision and inform a user on how to achieve a positive outcome. Our contribution in this paper is a framework called \textit{Counterfactual Generation with s(CASP) (CFGS)} that generates counterfactual explanations from \textit{rule-based machine learning (RBML)} algorithms. In doing so, we attempt to answer the question, ‘What can be done to achieve the desired outcome given that the outcome currently received is undesired?’. 

Our framework \textit{CFGS} models various 
worlds/scenarios---one is the current scenario where we are subject to the negative outcome and the other is the imagined scenario where we obtain the desired positive outcome. The idea is to move from the current world where the decision outcome is unfavorable to a world(s) where the desired outcome would hold given the static decision-making process (i.e., the decision maker does not change). The traversal between these worlds, i.e., from the original world where we got the negative outcome to the counterfactual world(s) where we get the positive decision, is done through \textit{interventions}, i.e., by changing the input \textit{feature values}. These interventions are made taking causal dependencies between features into account. Our framework relies on commonsense reasoning, as realized via answer set programming (ASP) \cite{gelfond-kahl}, specifically the goal-directed s(CASP) ASP system \cite{scasp-iclp2018}.


\vspace{-0.03in}
\section{Background}

\vspace{-0.05in}
\subsection{Counterfactual Reasoning}
As humans, we treat explanations as tools to help us understand decisions and inform us on how to act. Wachter et al. \cite{wachter} argued that counterfactual explanations (CFE) should be used to provide explanations for individual decisions. Counterfactual explanations offer meaningful explanations to understand a decision and inform on what can be done to change the outcome to a desired one. In the example of being denied a loan, counterfactual explanations are similar to the statement: If John was married, his loan application would have been approved. A key idea behind counterfactual explanations is imagining a different world where the desired outcome would hold. This different world ought to be reachable from the current world. Thus, the concept of ``closest" or ``close possible worlds" imagines alternate (reasonably plausible) scenarios where such a desired outcome would be achievable.

For a binary classifier used for prediction, given by $f:X \rightarrow \{0,1\}$, we define a set of counterfactual explanations $\hat{x}$ for a factual input $x \in X$ as $\textit{CF}_{f}(\hat{x})=\{\hat{x} \in X | f(x) \neq f(\hat{x})\}$. The set of counterfactual explanations contain all the inputs ($\hat{x}$) that lead to a prediction under  $f$ that is different from the prediction for the original input $x$. 

Additionally, Ustun et al. \cite{ref_2_ustun} highlighted the importance of \textit{algorithmic recourse} for counterfactual reasoning which we define as \textit{the process of altering undesired outcomes by algorithms in favor of obtaining desired outcomes through certain actions}. Building on top of the work by Ustun et al. \cite{ref_2_ustun}, Karimi et al. \cite{ref_3_karimi_1} showed in their work how to generate counterfactual explanations that take immutability of features into account. For example, a counterfactual instance that recommends changing the `gender' or `age' of an individual has limited utility. Their work allows restriction on the kind of changes that can be made to feature values in the process of generating plausible counterfactual explanations. Additionally, work of both Ustun et al \cite{ref_2_ustun} and Karimi et al \cite{ref_3_karimi_1} generates a set of diverse counterfactual explanations to choose from. However, they assume that features are independent of each other. In the real world, there may be causal dependencies between features. 

We show how counterfactual reasoning can be performed using the s(CASP) query-driven predicate ASP system \cite{scasp-iclp2018} \textit{while taking causal dependency between features into account}. By utilizing s(CASP)'s inbuilt ability to compute the \textit{dual rules} (as described in Section \ref{dual_rules}) that allow us to execute negated queries, counterfactual explanations can be naturally obtained. Given the definition of a predicate {\tt p} as a rule in ASP, its corresponding \textit{dual rule} allows us to prove {\tt not p}, where {\tt not} represents \textit{negation as failure} \cite{ref_NAF_1}. We utilize these \textit{dual rules} to construct alternate worlds that lead to counterfactual explanations, while taking causal dependencies between features into account. 

\vspace{-0.075in}
\subsection{Causality}

Taking inspiration from the Structural Causal model approach by Pearl et al. \cite{SCM}, Karimi et al.\cite{ref_4_karimi_2} showed how only considering the nearest counterfactual explanations in terms of cost/distance, without taking into account causal relations that model the world, produce unrealistic explanations that are not realizable. Their work focused on generating counterfactual explanation through a series of interventions that are readily achievable and that provide a realistic path to flipping the predicted label. In earlier approaches of generating counterfactuals \cite{ref_2_ustun} \cite{ref_3_karimi_1}, implicit assumptions are made where changes resulting from  interventions will be independent of changes across features. However, this is only true in worlds where features are independent. This assumption of independence across features might need to be revised in the real world. Causal Relationships that govern the world should be taken into account. For example, consider again the case where an individual wants to be approved for a loan. Given that the loan approval system takes into account the marital status, the counterfactual generation system which takes into account the marital status, the relationship status, the gender of the individual and the age, might make a recommendation of changing the marital status of the individual to `single'. However, the individual still has a relationship status of `husband'. A causal dependency exists between at least two features (marital status and relationship). In addition, the `gender' of the individual influences if the person is a `husband' or `wife'. Hence, a causal dependency exists between `gender' and `relationship'. Unless these causal dependencies are taken into account, a counterfactual explanation of changing one's marital status by assuming that relationship and gender are independent might be unrealistic and might not even result in the loan being approved (assuming the loan approval algorithm also looks at relationship status in addition to marital status). By highlighting the importance of modeling causal relationships which, in turn, model down-steam changes caused by directly changing features, realistic counterfactual explanations can be generated. 

\vspace{-0.1in}

\subsection{ASP, s(CASP) and Common Sense Reasoning}\label{dual_rules}

Answer Set Programming (ASP) is a well established paradigm for knowledge representation and reasoning \cite{cacm-asp,baral,gelfond-kahl} with many prominent applications, especially to automating commonsense reasoning. We use ASP to represent the knowledge of the features---the domains, the properties of the features, the decision making rules and the causal rules. We then utilize this encoded symbolic knowledge to automatically generate counterfactual explanations.

s(CASP) is a goal-directed ASP system that executes answer set programs in a top-down manner without grounding them \cite{scasp-iclp2018,ref_GG}. The query-driven nature of s(CASP) greatly facilitates performing commonsense reasoning as well as counterfactual reasoning based on commonsense knowledge employed by humans . Additionally, s(CASP) justifies counterfactual explanations by utilizing proof trees. 
 
In s(CASP), to ensure that facts are true only as a result of following rules and not by any spurious correlations, \textit{program completion} was adopted, which replaces a set of "if" rules with "if and only if" rules. One way to implement this was for every rule that says $p \implies q$, we add a complementary rule saying $\neg p \implies \neg q$. The effect ensures that q is true "if and only if" p is true.
In s(CASP), \textit{program completion} is done by introducing \textbf{dual rules}. For every rule of the form $p \implies q$, s(CASP) automatically generates dual rules of the form $\neg p \implies \neg q$ to ensure \textit{program completion}. 
Commonsense knowledge in ASP can be emulated using (i) default rules, (ii) integrity constraints, and (iii) multiple possible worlds~\cite{ref_GG,gelfond-kahl}.  We assume the reader is familiar with ASP and s(CASP). An introduction to ASP can be found in Gelfond and Kahl's book \cite{gelfond-kahl}, while a fairly detailed overview of the s(CASP) system can be found elsewhere \cite{arias-ec2022,scasp-iclp2018}.

\vspace{-0.1in}
\subsection{FOLD-SE}
Wang and Gupta \cite{foldse} devised an efficient, explainable, \textit{RBML} for classification tasks known as FOLD-SE. It comes from the FOLD (First Order Learner of Defaults) family of algorithms  \cite{fold,foldrpp,foldrm}. For given input data (numerical and categorical), FOLD-SE generates a set of default rules—essentially a stratified normal logic program—as an (explainable) trained model. The explainability obtained through FOLD-SE is scalable. Regardless of the size of the dataset, the number of learned rules and learned literals stay relatively small while retaining good accuracy in classification when compared to other \textit{RBML} approaches such as RIPPER \cite{ripper}. The rules serve as the model, and their accuracy is comparable with traditional tools such as XGBoost \cite{xgboost} and Multi-Layer Perceptrons (MLP) with the added advantage of being explainable. 

\section{Motivation}

We will explain our motivation through the following scenario: Suppose we apply for a bank loan and our application is rejected by an automated system that employs machine learning technology. Then, we would like to know (i) why such a decision was made and (ii) what options are available to us to obtain the given loan. This is due to our desire to see how to improve our chances of obtaining the loan after rejection. In this paper, we shall focus on (ii) the latter task of exploring what can be done to flip a negative outcome into a positive one. 
This task will involve understanding the causes of failure and curing them. The \textit{Counterfactual Generation with s(CASP) (CFGS)} framework's approach is simple: if decision-making rules (generated from \textit{RBML} algorithms) produce an undesired outcome, assuming that we are aware of the rules that have made the decision or prediction, we negate the query used for making the decision. This negated query must succeed. The proof tree for the negated query will tell us the changes/interventions we must make in order to force the negated query to fail and the original query to succeed. 

We will use answer set programming (ASP) to represent the rules and execute the query. ASP supports negation-as-failure \cite{ref_NAF_1,baral} thus allowing us to craft the negated query with ease. ASP is a well-established logic programming paradigm for knowledge representation and reasoning \cite{gelfond-kahl,cacm-asp} that supports sound semantics for \textit{negation}.

Let us consider a simple contrived example for illustration. Suppose we have a \textit{RBML} model that approves a loan based on a single feature of `marital status.' The loan is approved if the applicant is \textit{married}. Otherwise, if the applicant is \textit{single}, the application is rejected. This decision making can be captured by the rules shown below (in Prolog/ASP syntax):

{\tt Rule 1: approve\_loan(X) :- married(X).}   

\noindent 
Suppose John is a single person who applies for a loan. The loan application will obviously fail as the query 

\texttt{Query 1: ?-approve\_loan(john).} 

\noindent will fail. If the applicant wants to flip the decision, he/she should change the relevant feature(s) so that the query will succeed. To find out which feature(s) to change to cure the failure, we construct the negated query 

\texttt{Query 2: ?- not approved\_loan(john).} 

\noindent and execute it. The negated goals will be executed against the \textit{dual rule(s)} computed by the ASP engine:

{\tt Rule 2: not approved\_loan(X) :- not married(X).}  

\noindent The negated query (Query 2)  will surely succeed since the original query (Query 1) failed. The proof tree constructed for the successful negated query will indicate the feature that must be changed so that John can get a loan. Our aim is to intervene and change the feature values so that the negated query (Query 2) will fail (and, therefore, the original query (Query 1) will succeed). An examination of the simple proof tree of the negated query (Query 2) tells us that it succeeded because the subsequent goal {\tt not married(john)} succeeded. If John's status changes to {\tt married}, that will thwart the goal {\tt not married(john)}, and the negated query (Query 2) will fail. Therefore, The original query (Query 1) will succeed, and we can advise John that he should get married to secure a loan.


In the process of flipping a negative decision to a positive decision, our \textit{CFGS} framework will model 2 distinct scenarios---one where the loan is denied, which we shall refer to as a \textit{pre-intervention world}, and alternative imaginary scenario(s)/world(s) (where, after making interventions, we obtain the loan) that we refer to as a \textit{post-intervention world}. These two distinct scenarios/worlds are characterized by the decisions made by the decision-making system (the loan approval system). Recall that, in the loan approval example, we are denied a loan in the \textit{pre-intervention world}, while in the imagined \textit{post-intervention world}, we obtain the loan. So, the  decision in the \textit{pre-intervention world} should not hold in the \textit{post-intervention world}. 
Thus, the query original query {\tt ?- approve\_loan(john)} (Query 1) is {\tt False} in the \textit{pre-intervention world} and {\tt True} in the post-intervention world. 
Through our proposed framework \textit{CFGS}, we aim to sybmolically compute the interventions needed to flip a decision. The interventions can be thought of as finding a counterfactual explanation leading to the flipped decision (if John were married, the loan application would be approved). We use the s(CASP) query-driven predicate ASP system \cite{scasp-iclp2018} to achieve this goal.
The advantage of s(CASP) lies in automatically generating dual rules, i.e., rules that let us constructively execute negated queries (e.g., {\tt not approve\_loan/1} above).


\begin{figure}[htp]
    \centering
    \includegraphics[width=15cm]{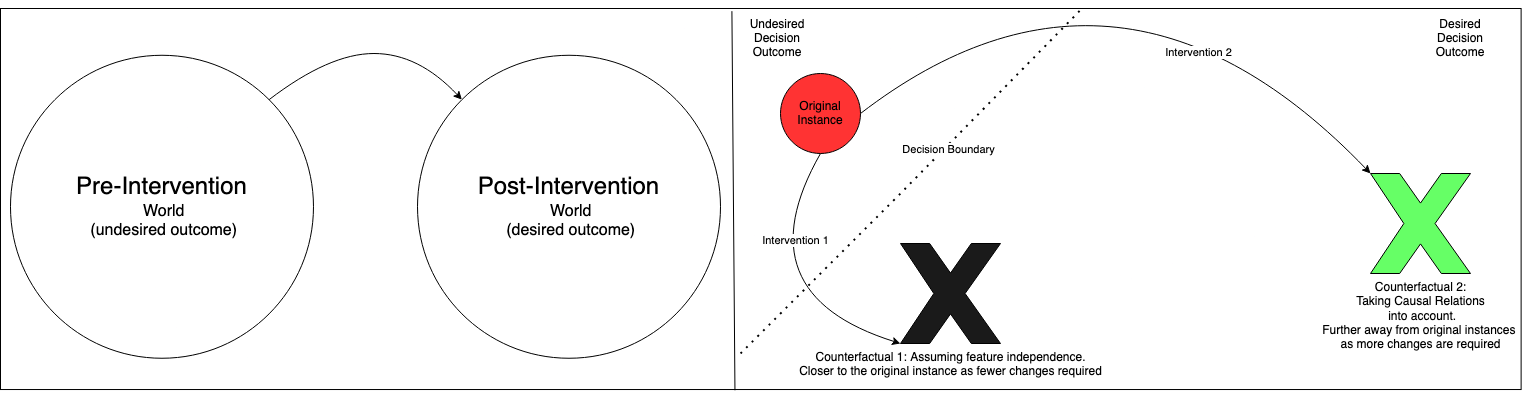}
    \caption{\textbf{Left:} Transition from \textit{Pre-Intervention World} to the \textit{Post-Intervention World. \textbf{Right:} Intervention takes the original instance to the other side of the decision boundary. With feature independence, the new counterfactual is closer to the original instance. With causal dependencies, the new counterfactual is further away as more changes are made to the original instance.}}
    \label{fig_1}
    \vspace{-0.2in}
\end{figure}


\vspace{-0.05in}

\section{Methodology}\label{Methodology}

In this section we define the methodology employed by our proposed framework \textit{CFGS} for generating counterfactual instances (that produce a \textit{desired outcome}) from an original instance (that produces an \textit{undesired outcome}). While we shall use a simple example to denote our process, we have run experiments (as show in Section \ref{Experiments}) on the following datasets: adult\cite{adult}, car\cite{car}, titanic\cite{titanic}, dropout\cite{dropout}, mushroom \cite{mushroom} and voting\cite{voting}. 

\begin{example}\label{Methodology_example}
Consider an example where a decision making \textit{RBML} algorithm produces rules. These rules classify if an individual defined by the features --- `relationship', `gender' and `age'--- is `married'. A \textit{RBML} algorithm generates the following rules that determine if an individual is married.: 
 
{\tt marital\_status(married) :- relationship(husband).} 

{\tt marital\_status(married) :- relationship(wife).} 
\end{example}

\begin{figure}[htp]
    \centering
    \includegraphics[width=14cm]{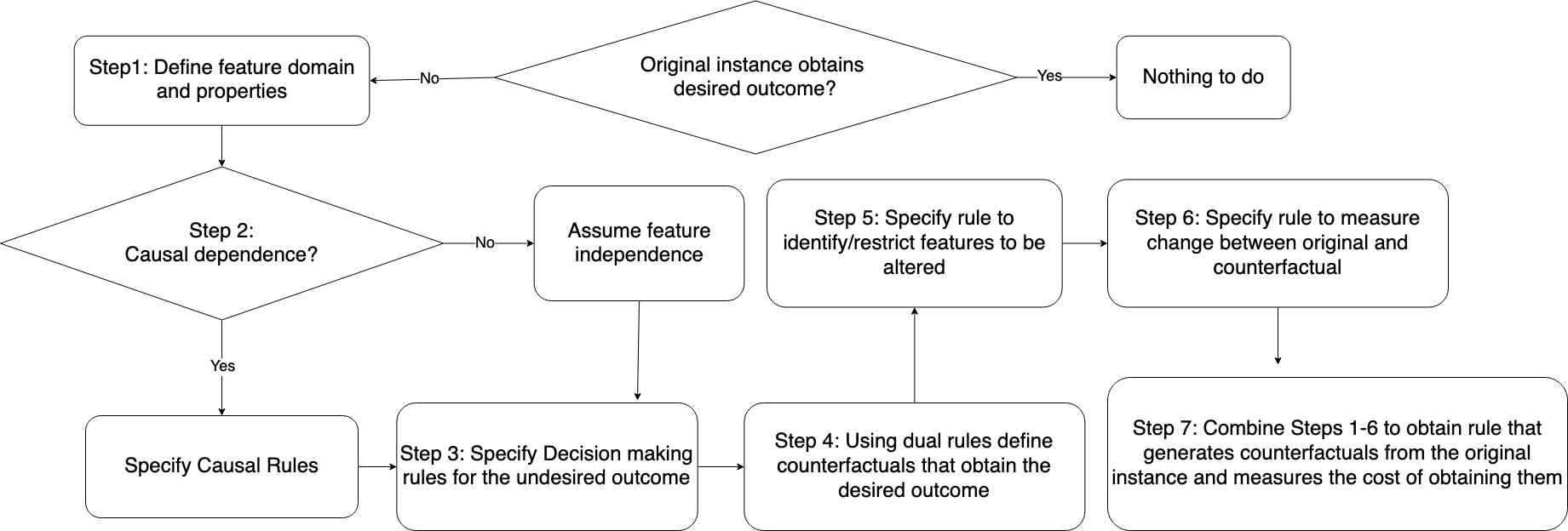}
    \caption{Methodology of the \textit{Counterfactual Generation with s(CASP) (CFGS)}}
    \label{fig_2}
\end{figure}

\noindent The above rules shown in Example \ref{Methodology_example} classify `married' individuals through their value of the `relationship' feature, i.e. `husband', `wife'. For an individual classified as `married', assuming that they do not wish to be classified as such, i.e., it is an undesired outcome, what would they have to do to not be classified as `married' i.e. be classified as `single'?










\subsection{Step 1: Defining Domain of Features}\label{Step_1}
We aim to determine what changes to make to obtain the desired outcome. Considering our running example \ref{Methodology_example}, what changes would lead an individual not to be classified as ‘married’ given that they were initially classified as ‘married’? In order to do that, we need to be aware of the details of the features, including their properties, such as if they are single-valued or multi-valued (a person can be either male or female but not both simultaneously) and the domain of values that the features can take. Which, in turn, determines what new values a feature can take if it is being changed.
\subsubsection{Defining Domains}
We determine the various features and feature values and represent them as facts. These facts essentially declare the domain values that each feature can take. Following example \ref{Methodology_example}, for categorical features, we represent the domains as follows:

{\tt 1. f\_domain(relationship,husband).}  

{\tt 2. f\_domain(relationship,wife).} 

{\tt 3. f\_domain(relationship, unmarried).}

{\tt 4. f\_domain(gender,male).}  

{\tt 5. f\_domain(gender,female).} 

\noindent The above facts represent the various values that the categorical feature can take. While `relationship' has values such as `husband,' `wife,' and `unmarried,' `gender' has values `male' and `female.'

For numerical features, we define the range of values that they can take. We can obtain them from the training data or specify them using common knowledge. 

{\tt 6. f\_domain(age, X) :- X \#>= 17, X \#=< 90.}

\noindent The above rule defines $17-90$ as the range of values that the numerical feature `age' can take.
After defining the domain of the features, the next step involves specifying their properties.

\subsubsection{Specifying Feature Properties}

We have two specific scenarios, one before change (pre-intervention), where we obtain an undesired outcome. The second scenario would be after change (post-intervention), where we make changes to obtain the desired outcome. We define our two scenarios/worlds as follows:
\begin{enumerate}
    \item \textbf{Pre-intervention:} The original scenario/world where we obtain an undesired outcome. In our example, it is the reality where the the individual is classified as `married'.
    \item \textbf{Post-intervention:} The imagined scenario/world that we aim to achieve. It is where we obtain the desired outcome. In our example, it is the imagined reality where the the individual is classified as `not married'.
\end{enumerate}
We need to specify the properties corresponding to each scenario to generate pre and post-intervention scenarios. 

\noindent \textbf{Numerical Features:} For numeric features, the properties of the features are expressed in the following rules, which specify the properties of the pre- and post-intervention world for the numerical feature `age.'

{\tt 7. pre\_age(X) :- f\_domain(age, X).}   

{\tt 8. post\_age(X) :- f\_domain(age, X).} 

\noindent 
For numerical features, it is implicit that each value is distinct, and two distinct values cannot exist simultaneously, i.e., if a feature value is 1, it cannot be 2 and vice versa. Hence, the property of mutual exclusivity is implicit for numerical features and does not have to be defined separately (numerical features are single-valued). 

\noindent\textbf{Categorical Features:} For categorical features, the properties of the features are expressed in the following rules. 

\label{rule_prop_pre}{\tt 9. not\_pre\_relationship(X) :- f\_domain(relationship, Y), 
pre\_relationship(Y), Y \textbackslash$=$ X.}  

{\tt 10. pre\_relationship(X) :- not not\_pre\_relationship(X).} 

\noindent 
The above rules specify the properties of the pre-intervention world for the feature `relationship.' We can similarly specify the pre-intervention properties for the feature `gender.'

\label{rule_prop_post}{\tt 11. not\_post\_relationship(X) :- f\_domain(relationship,Y), post\_relationship(Y), Y \textbackslash$=$ X.}  

{\tt 12. post\_relationship(X) :- not not\_post\_relationship(X).} 

\noindent 
The above rules specify the properties of the post-intervention world for the feature `relationship.' We can similarly specify the post-intervention properties for the feature `gender.'

Unlike numerical features, we need to specify the properties of mutual exclusivity for categorical features. As per our running example \ref{Methodology_example}, we know from common knowledge that a person is a husband or a wife. They cannot be both husband and wife simultaneously. Unfortunately, unless this mutual exclusivity is defined, s(CASP) will assume unrealistic scenarios where a person being both husband and wife simultaneously is True. Hence, for single-valued features, we need to encode this property for categorical features. The property of mutual exclusivity is encoded into the rules determining the pre-and post-intervention worlds.  It states that \textit{the feature `relationship' can take any value in the domain of feature `relationship' as long as it is not any other value in the domain of feature `relationship.'}

\vspace{-0.05in}

\subsection{Step 2: Incorporating Causal Rules for Modeling Realistic Solutions}\label{Step_4}

In the real world, it is seldom true that the features used are independent. More often than not, a causal dependency exists among features. In such scenarios, if there are causal dependencies amongst features, we must model them to produce realistic solutions. Considering our running example \ref{Methodology_example}, consider two features, ‘relationship’ and ‘gender.’ Suppose we assume that \textbf{no causal dependency} exists among them. In that case, i.e., they are independent, there can be instances where an individual has the value of ‘husband’ for the feature ‘relationship’ while having the value ‘female’ for the feature ‘gender.’ However, this is unrealistic. Hence, to model real-world scenarios and produce realistic counterfactual solutions, we incorporate causal relations into our algorithm.

\vspace{-0.05in}
\subsubsection{Identifying and Defining Causal Rules}\label{Step_2_4_1}

Using knowledge from domain experts or common sense knowledge, we identify features with cause-effect relations. We then run \textit{RBML} algorithms to obtain rules defining the causal relations we can incorporate into our algorithm. For example \ref{Methodology_example}, we obtain rules corresponding to the cause-effect relations between the features ‘gender’ and ‘relationship’ and express them in a format compatible with our framework \textit{CFGS}.

{\tt 13. causal\_relationship\_gender(husband,Y) :- Y \textbackslash$=$ female.} 

\noindent The above rule in line $13$ has two arguments with the first corresponding to the feature `relationship' and the second corresponding to the feature `gender.' This rule is \textbf{True} if the second argument $Y$ (corresponding to gender) is \textbf{not} female. In other words, the rule highlights the causal dependency between the `relationship' value `husband' and the feature `gender.' It says that a `husband' cannot be `female.'

\vspace{-0.05in}
\subsubsection{Using External Knowledge in the Absence of Rules}\label{Step_2_4_2}


The rule in line $13$ captured by our \textit{RBML} algorithm defines the causal dependency of feature `relationship' having value `husband' on the `gender' of the individual. However, we must also incorporate cases where certain causally dependent features do not generate rules for specific feature values. If we do not define rules for these feature values, our system will \textbf{never} produce/generate the corresponding feature values in the solutions. For example, the rule in line $13$ only defines the causal dependency of the `relationship' value 'husband' on the `gender' of the individual. Unless we define a causal rule for the `relationship' value of `wife,' our algorithm will \textbf{never} generate any instance with a `relationship' value of `wife.' To avoid this, in the scenario where our algorithm cannot generate rules, we use common knowledge or incorporate knowledge from a domain expert to add rules for cases/feature values for which our \textit{RBML} algorithm did not capture rules. 

{\tt 14. causal\_relationship\_gender(wife,Y) :- Y \textbackslash$=$ male.} 


\noindent The above rule in line $14$ has two arguments, corresponding to the features ‘relationship’ and ‘gender' respectively. This rule is \textbf{True} if the second argument $Y$ (corresponding to gender) is \textbf{not} ' male.’ In other words, the rule highlights the causal dependency of the feature ‘relationship‘ having a value of ‘wife’ on the feature ‘gender.’ It says that a ‘wife’ cannot be ‘male’. The above rule defines the causal dependency of the feature ‘relationship’ having a value of ‘wife’ from common knowledge. We attempt to define rules for all causal dependencies if they do exist. For example \ref{Methodology_example}, we have captured all causal dependencies in the form of the rules defined in Section \ref{Step_2_4_1} and Section \ref{Step_2_4_2}.

\vspace{-0.1in}
\subsubsection{Incorporating Causality for a Realistic Solution}
\label{Step_2_4_3}

Since we wish to model realistic solutions, we must take into account the causal rules defined in Section \ref{Step_2_4_1} and Section\ref{Step_2_4_2}. By combining these rules with the feature domains and feature properties defined in \ref{Step_1}, we define rules that generate realistic instances that follow the causal rules that govern the world. Taking our running example \ref{Methodology_example}, we define the following rules:

{\tt 15. pre\_realistic(X,Y,Z) :- 
f\_domain(relationship,X), f\_domain(gender,Y), f\_domain(age,Z),

\indent\indent\indent\indent\indent\indent\indent\indent\indent\indent\indent\indent\indent\indent pre\_relationship(X), pre\_gender(Y), pre\_age(Z),

\indent\indent\indent\indent\indent\indent\indent\indent\indent\indent\indent\indent\indent\indent causal\_relationship\_gender(X,Y).
} 

\noindent The above rule generates realistic instances for the pre-intervention scenario. Similarly, after making changes to obtain a counterfactual instance, we wish for such instances to be realistic, so we define a rule to generate realistic instances post-intervention. This is expressed as follows:

{\tt 16. post\_realistic(X,Y,Z) :- 
f\_domain(relationship,X), f\_domain(gender,Y), f\_domain(age,Z),

\indent\indent\indent\indent\indent\indent\indent\indent\indent\indent\indent\indent\indent\indent post\_relationship(X), post\_gender(Y), post\_age(Z),

\indent\indent\indent\indent\indent\indent\indent\indent\indent\indent\indent\indent\indent\indent causal\_relationship\_gender(X,Y).
} 

\noindent
By incorporating the rules in lines $15$ and $16$ into our overall function as we shall do later, we constrain our solutions to be realistic in nature and model the reality of the world that we live in.

\vspace{-0.1in}
\subsubsection{Causally Independent Features}

Assuming that no causal dependency exists amongst the features, then we slightly modify the rules from lines $15-16$ from Section \ref{Step_2_4_3}

{\tt 15. pre\_realistic(X,Y,Z) :- 
f\_domain(relationship,X), f\_domain(gender,Y), f\_domain(age,Z),

\indent\indent\indent\indent\indent\indent\indent\indent\indent\indent\indent\indent\indent\indent pre\_relationship(X), pre\_gender(Y), pre\_age(Z).
} 

\noindent The above rule generates the instances for the pre-intervention scenario where the features are independent and are not constrained by causal dependencies. 

{\tt 16. post\_realistic(X,Y,Z) :- 
f\_domain(relationship,X), f\_domain(gender,Y), f\_domain(age,Z),

\indent\indent\indent\indent\indent\indent\indent\indent\indent\indent\indent\indent\indent\indent post\_relationship(X), post\_gender(Y), post\_age(Z).
} 

\noindent 
The above rule generates the instances for the post-intervention scenario where the features are independent and none of the features are constrained by causal dependencies. 

\vspace{-0.05in}
\subsection{Step 3: Decision-making Rules for Undesired Outcomes}\label{Step_2}


Recall that in example \ref{Methodology_example}, we run an \textit{RBML} algorithm to obtain the original decision-making rules that produced the undesired outcome and express it in the ASP/s(CASP) syntax as follows:

\label{rule}
{\tt marital\_status(married) :- relationship(husband).}

{\tt marital\_status(married) :- relationship(wife).} 

\noindent 
The above rules state that a person is classified as `married' if the feature `relationship' has value `husband' or `wife'. However in order to work in our \textit{CFGS} framework, we need to rewrite these rules as follows:

\label{rule_pre_scasp}{\tt 17. lite\_married(X) :- X = husband.} 

{\tt 18. lite\_married(X) :- X = wife.} 

\noindent 
The rules in lines $17-18$ specify the decision-making component of our overall rule (as defined later in line 19) that determines if an individual is ‘married’ per our original decision-making rule in example \ref{Methodology_example}. The rule in line $17$ will be True if the variable $X$ has value  `husband.' Similarly, the rule in line $18$ will be True if the variable $X$ has value `wife.' 
Now we need to incorporate the domain and properties of the features as mentioned in Step 1 (Section \ref{Step_1}) and 2 (Section \ref{Step_4}). Our overall rule with the domain and feature properties defined for the pre-intervention scenario (before intervention is done to generate the counterfactual) is defined as follows:


{\tt 19. married(A) :- f\_domain(relationship,A), pre\_relationship(A), lite\_married(A).}


\noindent By querying the rules described in line 15 and 19 and leaving the variables unassigned, we can obtain the feature values for all the possible individuals that are classified as `married'.

{\tt ?- married(A), pre\_realistic(A,B,C).} 


\noindent In the query shown above, $A$ is the variable that represents the feature `relationship,' and $B$ is the variable that represents the feature `gender.' $C$ is the variable that represents the feature `age.' By leaving the variables $A$, $B$, and $C$ unassigned, the executed query above gives a symbolic representation of the solution space for all individuals classified as `married' (undesired outcome). The second component {\tt pre\_realistic(A,B,C)} generates realistic instances while the first component {\tt married(A)} constrains these realistic instances to achieve the original \textit{undesired} decision outcome.


\vspace{-0.05in}
\subsection{Step 4:  Counterfactuals in s(CASP)}\label{Step_3}

For example \ref{Methodology_example}, we aim to obtain the counterfactual instances for the decision-making rules that classify the individual with the undesired outcome of ‘married,’ i.e., we define rules that classify individuals that are ‘not married.’

\vspace{-0.05in}
\subsubsection{Procedure to Obtain Counterfactual Rules 
}
\label{Step_4_Procedure}
Our procedure for defining the counterfactual rules is as follows:
\begin{enumerate}
    \item Define the domain of the features as per the original decision making rules. 

    \item Define the features' properties per the original decision-making rules but only change pre-intervention property to post-intervention property (By definition, counterfactual instances are ones that we obtain \textbf{post/after} making an intervention). For example, if the property of the feature ’relationship’ in the original decision-making rules is given by {\tt pre\_relationship(A)}, we define the properties of the counterfactual instance as {\tt post\_relationship(A)} in the counterfactual formula.

    \item Negate the decision-making component of the original decision-making rules. For example, in the original decision-making rule in line $19$, the predicate {\tt lite\_married(X)} defines the decision-making component. We replace {\tt lite\_married(X)} with {\tt \textbf{not} lite\_married(X)} for the counterfactual rule.
\end{enumerate}

\vspace{-0.05in}
\subsubsection{Overall Counterfactual Rule:}

For our running example \ref{Methodology_example}, by following the procedure in
\ref{Step_4_Procedure}, we express the counterfactual rule that identifies counterfactual instances that disagree with the original decision making rule: 

{\tt 20. cf\_married(A1) :- f\_domain(relationship,A1), post\_relationship(A1), \textbf{not} lite\_married(A1).} 

\noindent
By querying the rules described in line 16 and 20 and leaving the variables unassigned, we can obtain the feature values for all the possible individuals that are \textbf{not} classified as ‘married’ (for example: single, divorced).




{\tt ?- cf\_married(A1), post\_realistic(A1,B1,C1).} 

\noindent In the query shown above, $A1$ is the variable that represents the feature `relationship', $B1$ is the variable that represents the feature `gender' and  $C1$ is the variable that represents the feature `age'. By leaving the variables $A1$, $B1$ and $C1$ unassigned, the executed query above gives a symbolic representation of the counterfactual solution space for all individuals that will be \textbf{not} be classified as `married' (for example: single, divorced). The second component {\tt post\_realistic(A1,B1,C1)} generates realistic instances while the first component {\tt cf\_married(A1)} constrains these realistic instances to be counterfactual instances that achieve the \textit{desired} decision outcome.

\vspace{-0.05in}

\subsection{Step 5: Restricting Features - Taking Mutability and Immutability of Features into Account}\label{Step_2_5_restrict}

Our goal is to suggest interventions on the original instance (that produces an undesired outcome) that will lead to a counterfactual instance (that produces the desired outcome). In many cases, it will not be possible for individuals to change certain feature values. For our running example \ref{Methodology_example}, take the case of the feature `gender'. A counterfactual that recommends changing the value of the categorical feature `gender' from `male' to `female' and vice versa will be unrealistic. 
Additionally, there might be scenarios where the counterfactual might make other suggestions such as reducing the value of the numeric feature `age' which is also unrealistic. Hence to tackle this problem, we introduce rules that \textbf{constrain/restrict} the kind of changes (interventions) that can be made to the features in the process of generating counterfactuals.
If a certain feature is declared as immutable, its value cannot be changed. On the other hand, if a feature is mutable, it is allowed to change its value. If no such declaration is made for a feature, it can be treated as a mutable feature, if the need arises.



\vspace{-0.05in}
\subsubsection{Constraining/Restricting Categorical Features } \label{Step_2_5_1}

We aim to define the rules that constrain/restrict the kind of interventions/changes that can be made to categorical features.

{\tt 21. f\_domain(restrict\_C,0).} \indent \indent \indent \indent \indent 
{\tt f\_domain(restrict\_C,1).} 

\noindent The facts above define kind of interventions that restrict categorical features. If it equals $0$, the feature is immutable and if it equals $1$ then the feature's value has to be altered/changed/intervened on. 

{\tt 22. compare\_C(Pre\_X,Post\_X,Z) :- f\_domain(restrict\_C,Z), Z = 0, Pre\_X = Post\_X. 

23. compare\_C(Pre\_X,Post\_X,Z) :- f\_domain(restrict\_C,Z), Z = 1, Pre\_X \textbackslash$=$ Post\_X.} 

\noindent 
The above rules (lines $22$-$23$) restrict the kind of pre- and post-intervention on categorical features. If $Z = 0$ the feature remains unchanged or is immutable before and after intervention. If $Z = 1$ the pre-intervention value of the feature changes post-intervention, i.e., the feature values changes after intervention.

\vspace{-0.05in}
\subsubsection{Constraining/Restricting Numeric Features } \label{Step_2_5_2}
We aim to define the rules that constrain/restrict the kind of interventions/changes that can be made to numeric features.

{\tt 24. f\_domain(restrict\_N,0).} \indent \indent \indent 
{\tt f\_domain(restrict\_N,1).}  \indent \indent \indent 
{\tt f\_domain(restrict\_N,-1).} 

\noindent The facts above define kind of interventions that restrict numeric features. $0$ indicates a feature is immutable.$1$ indicates that the value of the numeric feature will increase and $-1$ indicates that the value will decrease after intervention


{\tt 25. compare\_N(Pre\_X,Post\_X,Z) :- f\_domain(restrict\_N,Z), Z = 0, Pre\_X = Post\_X. 

26. compare\_N(Pre\_X,Post\_X,Z) :- f\_domain(restrict\_N,Z), Z = 1, Pre\_X \#< Post\_X. 

27. compare\_N(Pre\_X,Post\_X,Z) :- f\_domain(restrict\_N,Z), Z = -1, Pre\_X \#> Post\_X.} 

\noindent 
The above rules in lines $25-27$ restrict the kind of intervention on numeric features, pre and post  intervention (before and after making an intervention). If $Z = 0$ the feature is immutable or does not change pre and post intervention. If $Z = 1$ the post intervention value of the feature is higher than the pre-intervention value, i.e., the feature value increases after intervention. If $Z = -1$ the post intervention value of the feature is lower than the pre-intervention value, i.e., the feature value decreases after intervention. 

\vspace{-0.05in}




\subsubsection{Overall Rule Restricting Interventions Between the Original and Counterfactual Instances} \label{Step_2_5_3}

Taking our running example \ref{Methodology_example} into account, we define an overall rule for restricting the interventions between an original instance and counterfactual instances by using the rules defined in \ref{Step_2_5_1} and \ref{Step_2_5_2}

{\tt 28. id\_restrict(original(X1,X2,X3), id(Z1,Z2,Z3), counterfactual(Y1,Y2,Y3)) :-

\indent \indent \indent \indent \indent \indent \indent \indent \indent \indent \indent \indent compare\_C(X1,Y1,Z1), compare\_C(X2,Y2,Z2), 
compare\_N(X3,Y3,Z3).}

\noindent The above rule in line $28$ takes in 3 arguments; $original(X1,X2,X3)$ which represents the original instance with three features $X1$, $X2$ and $X3$ before any intervention, $counterfactual(Y1,Y2,Y3)$ which represents the counterfactual instance with three features $Y1$, $Y2$ and $Y3$ generated after intervention and $id(Z1,Z2,Z3)$ which represents the kind of restriction that is allowed on the features. $Z1$ indicates the restriction on the feature $X1$ as its value changes to $Y1$, $Z2$ indicates the restriction on the feature $X2$ as its value changes to $Y2$ and $Z3$ indicates the restriction on the feature $X3$ as its value changes to $Y3$. Additionally, by leaving any of the variables $Z1,Z2$ and $Z3$ unassigned, we put \textbf{no restriction} on the corresponding features, i.e., their values can change if needed.

\vspace{-0.075in}
\subsection{Measuring the Cost of Interventions}\label{Step_2_6}

To ensure that we do not make unnecessary interventions that require an investment of time and effort, we measure the cost of making interventions on the original instance in the process of generating counterfactuals. Our definition of the cost involves the sum total of all features that have been intervened on in the process of generating the counterfactual. Our approach to this is simple, we identify intervened categorical features (features that have been changed) by the value \{$1$\} and immutable  categorical features by the value \{$0$\} as shown in Section \ref{Step_2_5_1}. Similarly we identify intervened numeric features by the values \{$1$, $-1$\} and immutable numeric features by the value \{$0$\} as shown in Section \ref{Step_2_5_2}. We simply sum up the categorical values and sum of the squares of the numerical values and add both the sums to obtain the total cost. Hence the maximum value for the cost of intervention is equal to the total number of features.

\vspace{-0.05in}
\subsubsection{Step 6: Measuring the Cost of Changed Features}\label{Step_2_6_1}
As per our running example \ref{Methodology_example}, we define a rule for computing the total number of features that have been intervened which we measure as the cost of making interventions.

{\tt 29. measure(Z1,Z2,Z3,X) :- f\_domain(restrict\_C,Z1), 
f\_domain(restrict\_C,Z2), 

\indent\indent\indent\indent\indent\indent\indent\indent\indent\indent\indent f\_domain(restrict\_N,Z3),
Q3 \#= Z3*Z3, X \#= Z1+Z2+Q3.}

\noindent The above rule in line $29$ calculates the total cost of making changes to the original instance. Here {\tt Z1, Z2} and {\tt Z3} are the \textbf{restriction} variables that can also indicate whether a feature has changed after intervention. They have the domain \{0,1,\} for categorical features and \{0,1,-1\} for numerical features. For the above rule in line 29, {\tt Z1, Z2} indicate the restriction on the categorical features whereas {\tt Z3} indicates the restriction on the numeric feature. For categorical features, as mentioned earlier, simply summing up the values will give the total number of features that have been intervened on/changed. For numerical features the value \{-1\} indicates that the counterfactual feature value was decreased after intervention. So in order to measure the cost, we square the restriction variables for the numerical features. Hence 0 will remain 0, 1 will remain 1 and -1 will become 1. By summing them up, we can compute the total number of numeric features that have been intervened on. Adding the overall sums/cost for the numerical and categorical features provides the overall cost of making interventions (total number of features that have been intervened on).

\vspace{-0.05in}
\subsection{Step 7: Obtaining Counterfactual from the Original Instance}\label{Step_6}

As per our running example, we incorporate all the rules from Steps 1 through 6 into one final rule to obtain

{\tt 30. refined(original(A,B,C),id(Z1,Z2,Z3),counterfactual(A1,B1,C1),X) :- married(A,B,C),

\indent\indent\indent\indent\indent pre\_realistic(A,B,C),
	cf\_married(A1,B1,C1), post\_realistic(A1,B1,C1),
 
\indent\indent\indent\indent\indent id\_restrict(original(A,B,C), id(Z1,Z2,Z3), counterfactual(A1,B1,C1)),
 
\indent\indent\indent\indent\indent  measure(Z1,Z2,Z3,X).
}

\begin{enumerate}
    \item the first argument $original(A,B,C)$ represents the original feature values that led to the undesired decision,  
    \item the second argument $id(Z1,Z2,Z3)$ represents the constraints on each of the features, while 
    \item the third argument $counterfactual(A1,B1,C1)$ represents the counterfactual values, while  
    \item the fourth argument $X$ represents the cost of making the interventions. 
\end{enumerate}


{\tt ?- refined(original(A,B,C),id(Z1,Z2,Z3),counterfactual(A1,B1,C1),X).} 

\noindent By leaving the variables $A$, $B$, $C$, $Z1$, $Z2$, $Z3$, $A1$, $B1$, $C1$  unassigned, the executed query above gives a symbolic representation of the solution space for possible paths that can be taken by original instances (that were subject to an undesired outcome) in the process of generating counterfactual instances. It also specifies the cost that will be incurred by the original instance in taking a path to achieving a particular counterfactual.

\begin{figure}[htp]
    \centering
    \includegraphics[width=15cm]{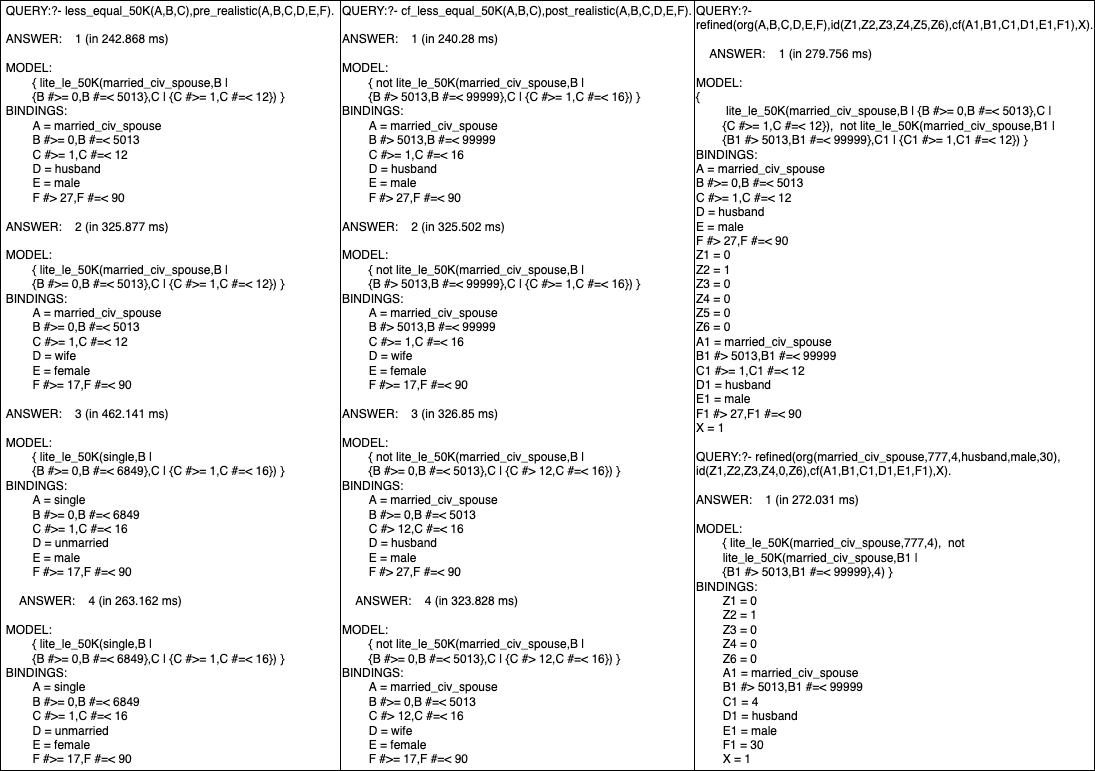}
    \caption{Query Results (adult): \textbf{Left:} We run the query with unassigned variables to obtain a list of instances that will be classified with the undesired result by the decision making rules. \textbf{Middle:} We run the counterfactual query with unassigned variables to obtain a list of counterfactual solutions for the given decision rules. \textbf{Right:} We first run the query to achieve the original-counterfactual pairs with unassigned variables to get all possible Original Instance- Counterfactual pairs with the cost highlighted. Secondly, we run the same query for a particular individual with an undesired outcome and obtain counterfactual solutions indicating what features need to be changed (here it was suggested to increase \textit{ the value of B from 777})}
    \label{fig_3}
\end{figure}

Additionally, by specifying the the respective feature values for the original instances, we can obtain a possible counterfactual solution.  Figure \ref{fig_3} shows the results of running this query on the adult dataset \cite{adult} that has six mixed features (3 categorical and 3 numerical). The bottom query in the third column shows an original instance and a corresponding counterfactual solution.
 Consider the query for our running example:

{\tt ?- refined(original(A,B,C),id(Z1,Z2,Z3),counterfactual(A1,B1,C1),1).} 

\noindent 
By specifying the value of the fourth argument/cost as we have shown in the query above, we can also restrict the solution space for all paths taken to have a specific cost (in this case, the cost is 1).



\vspace{-0.1in}
\section{Experiments}\label{Experiments}




For our running example \ref{Methodology_example}, we have defined a corresponding rule in Section \ref{Step_6} that gives us \textit{original instance-counterfactual} pairs that denoted the possible paths that can be taken by an original instance (subjected to a undesired outcome) in reaching a counterfactual (as shown in the rightmost column of figure \ref{fig_3} for the adult dataset). It also gives us  the cost incurred in reaching the counterfactuals while also allowing flexibility in what kind of intervention is permitted in the process of generating such counterfactuals. 

We have applied our \textit{CFGS} methodology to rules generated by \textit{RBML} algorithms FOLD-SE \cite{foldse} and RIPPER \cite{ripper}. The datasets that were used for generating the rules are as follows: adult\cite{adult}, car\cite{car}, titanic\cite{titanic}, dropout\cite{dropout}, mushroom \cite{mushroom} and voting\cite{voting}. 
The adult dataset that contains demographic information of individuals with the label indicating whether someone makes `$=<$\$50k/year', or `$>$\$50k/year'. 
In our experiments on the adult dataset, our \textit{RBML} algorithms give us rules indicating whether someone makes `$=<$\$50k/year'. Given that the decision making rules obtained from these datasets specify an undesired outcome (a person making less than or equal to \$50K) for an original instance, the goal is to find a path to a counterfactual instance where a person makes more than \$50K.

Our \textit{CFGS} methodology produces a list of \textit{original-counterfactual} pairs that denote all the possible paths to be taken by an original instance in reaching a counterfactual along with the cost incurred. We have shown our results in table \ref{tbl_Categorical_Numeric}.

 \begin{table}[h!]
 \vspace{-0.15in}
   \centering
   \resizebox{\columnwidth}{!}{%
   \renewcommand{\arraystretch}{2}
   
   \begin{tabular}{|p{2cm}|c|c|c|c|c|c|c|c|c|}
     \hline
     \multicolumn{1}{|c|}{\textbf{Dataset}} & \multicolumn{3}{c|}{Adult} & \multicolumn{3}{c|}{Titanic} & \multicolumn{3}{c|}{Dropout} \\
     \cline{1-10}
     \multicolumn{1}{|c|}{\textbf{Details}} & \# of Features Used & Training Size & Time Taken (ms) & \# of Features Used & Training Size & Time Taken (ms) & \# of Features Used & Training Size & Time Taken (ms) \\
     \hline
     \multicolumn{1}{|c|}{\textbf{FOLD-SE:}} & 6  & 26048 & 291  & 3 & 891 & 84  & 4 & 3539 & 67  \\ \hline
     \multicolumn{1}{|c|}{\textbf{RIPPER:}} & 12  & 26048 & 2869  & 1 & 891 & 14 & 16 & 3539 & 807  \\ \hline \hline \hline

      \hline
      \multicolumn{1}{|c|}{\textbf{Dataset}} & \multicolumn{3}{c|}{Voting} & \multicolumn{3}{c|}{Cars} & \multicolumn{3}{c|}{Mushroom} \\
      \cline{1-10}
      \multicolumn{1}{|c|}{\textbf{Details}} & \# of Features Used & Training Size & Time Taken (ms) & \# of Features Used & Training Size & Time Taken (ms) & \# of Features Used & Training Size & Time Taken (ms) \\
      \hline
      \multicolumn{1}{|c|}{\textbf{FOLD-SE:}} & 5  & 348 & 779  & 4 & 1382 & 807  & 5 & 6499 & 600  \\ \hline
      \multicolumn{1}{|c|}{\textbf{RIPPER:}} & 2  & 348 & 75  & 6 & 1382 & 75 & 9 & 6499 & 3911  \\ \hline

   \end{tabular} 
   }
   
   \vspace{0.05in}
   \caption{Time for Computing the Counterfactual for Various Datasets}
   \label{tbl_Categorical_Numeric}
   \vspace{-0.4in}
 \end{table}

For each dataset in the table \ref{tbl_Categorical_Numeric}, there are 3 columns denoting the following:
\begin{enumerate} 
    \item Number of Features: Count of the features that were used in generating \textit{original instance-counterfactual} pairs. This depends on the features defined in the decision making rules (Step \ref{Step_2}) and causal rules (Step \ref{Step_4}). 
    \item Size of Training Data: Size of the training data used to generate the decision making and causal rules.
    \item Time Taken: Time taken to produce the \textit{(original instance, counterfactual)} pair. 
    
\end{enumerate}

As shown in table\ref{tbl_Categorical_Numeric}, our \textit{CFGS} framework generates counterfactuals regardless of the \textit{RBML} algorithm that specifies the decision making rules.

\vspace{-0.1in}
\section{Discussion and Related Work}
There are existing approaches to tackle this problem of lack of transparency by providing an explanation to an undesired outcome such as that of Wachter et al. \cite{wachter}. Some approaches are tied to particular models or families of models, while some use optimization-based approaches to try and solve the same problem \cite{ref_mace_1,ref_mace_2}. Ustun et al. \cite{ref_2_ustun} took an approach that highlighted the need to focus on algorithmic recourse  to ensure a viable counterfactual explanation. Others have found ways to utilize counterfactual explanations to improve model performance and ensure accurate explanations such as White and Garcez \cite{ref_clear}. Lately, specific approaches have considered the context of consequential decision-making concerning the type of features being altered and focused on producing viable realistic counterfactuals such as the work by Karimi et al. \cite{ref_3_karimi_1} which have the additional advantage of being model-agnostic. Our framework \textit{Counterfactual Generation with s(CASP) (CFGS)} provides possible explanations that justify a decision through counterfactual explanations by utilizing the answer set programming paradigm (ASP). Additionally it has the flexibility to generate counterfactual instances regardless of the \textit{RBML} algorithm used while also allowing injection of user-defined rules. While there have been other approaches utilizing ASP such as Bertossi and Reyez \cite{ref_asp_cf}, it does not make use of a goal-directed ASP system and relies on grounding which has a disadvantage of losing the association between variables. Our approach utilizes s(CASP) and hence no grounding is required. Additionally we can implement complex logic over the control variables (Z1-Z6) that allows us to determine the mutability of certain features.
The main contribution of this paper is the CFGS framework that shows that complex tasks such as imagining possible scenarios, which is essential in generating counterfactuals as shown by Byrne \cite{ref_Byrne_CF} (``what if something else happened?"), can be modeled with s(CASP) while taking causal dependencies between features into account. Such imaginary worlds (where alternate facts are true and, hence, different decisions are made) can be automatically computed using the s(CASP) system. 
Detailed explanations are also provided by s(CASP)  for any given decision reached. The s(CASP) system thus allows for counterfactual situations to be imagined and reasoned about.

\vspace{-0.1in}
\section{Conclusion}

In this paper, we addressed the problem of generating counterfactual explanations for an undesired outcome predicted by a machine learning model. A major problem related to explainability is the problem of automatically determining changes that must be made to the input feature values in order to flip the decision from an undesired outcome to a desired one. Computing these changes amounts to finding counterfactual explanations. We have proposed a framework \textit{Counterfactual Generation with s(CASP) (CFGS)} that demonstrates how ASP and specifically the s(CASP) goal-directed ASP system can be used for generating these counterfactual explanations regardless of the \textit{RBML} algorithm used for generating the decision making rules. The s(CASP) system's support for negation as failure and dual rules allow us to generate alternate worlds. To generate counterfactual explanations, we need to find alternate worlds that are reachable from the current world. We showed in a methodical manner how to reach a counterfactual from a given instance. We additionally highlighted how to identify and constrain features to be altered in the process of generating counterfactual instances, in addition to measuring the cost of making such interventions. Our approach allows specifying the cost to be incurred in generating counterfactual explanations and also allows user-defined rule to be added in the absence of \textit{RBML} algorithm rules.

\bibliographystyle{ACM-Reference-Format}
\bibliography{sample-base}

\end{document}